\documentclass{article}

\usepackage{proceed2e}
\usepackage{grffile}
\usepackage{graphicx}  
\usepackage{amsmath}
\usepackage{tikz}
\usepackage{aircraftshapes}
\usepackage{subfig}
\usepackage{pgfplots}
\usepackage{pgfplotstable}
\usepackage{natbib}
\usepackage{booktabs}
\usepackage{hyperref}
\usepackage[capitalize]{cleveref}
\usepackage[per-mode=symbol,detect-all=true,group-separator={,},group-four-digits=false,group-decimal-digits=false,load=accepted,list-final-separator={, and }]{siunitx}

\newunit{\feet}{ft}
\newunit{\nauticalmile}{NM}
\newcommand{\argmax}{\operatornamewithlimits{arg\,max}}

\pgfplotsset{compat=newest}
\pgfplotsset{every axis legend/.append style={%
cells={anchor=west},
}}
\usetikzlibrary{arrows,fit,positioning,calc}
\usepgfplotslibrary{groupplots}

\hypersetup{
    colorlinks=true,
    hypertexnames=false,
    plainpages = false,
    breaklinks=true,
    a4paper=false,
    linkcolor = black,  
    anchorcolor = black,
    citecolor = black,
    filecolor = black,
    menucolor = black,
    urlcolor = black,
    pdfauthor={},
    pdftitle={Predicting the behavior of interacting humans by fusing data from multiple sources},
    pdfkeywords={}
}

\usepackage{color}  

\begin{document}
\title{Predicting the behavior of interacting humans\\ by fusing data from multiple sources} 
\author{ 
{\bf Erik J.~Schlicht$^1$, Ritchie Lee$^2$, David H. Wolpert$^{3,4}$,} \\{\bf  Mykel J. Kochenderfer$^1$, and  Brendan Tracey$^5$} \\
$^1$ Lincoln Laboratory, Massachusetts Institute of Technology, Lexington, MA 02420 \\
$^2$ Carnegie Mellon Silicon Valley, NASA Ames Research Park, Moffett Field, CA 94035 \\
$^3$ Information Sciences Group, MS B256, Los Alamos National Laboratory, Los Alamos, NM 87545 \\
$^4$ Santa Fe Institute, 1399 Hyde Park Road, Santa Fe, NM 87501 \\
$^5$ Stanford University, Palo Alto, CA 94304 
}
\maketitle 
\begin{abstract} 
Multi-fidelity methods combine inexpensive low-fidelity simulations with costly but high-fidelity simulations to produce an accurate model of a system of interest at minimal cost. They have proven useful in modeling physical systems and have been applied to engineering problems such as wing-design optimization. During human-in-the-loop experimentation, it has become increasingly common to use online platforms, like Mechanical Turk, to run low-fidelity experiments to gather human performance data in an efficient manner. One concern with these experiments is that the results obtained from the online environment generalize poorly to the actual domain of interest. To address this limitation, we extend  traditional multi-fidelity approaches to allow us to combine fewer data points from high-fidelity human-in-the-loop experiments with plentiful but less accurate data from low-fidelity experiments to produce accurate models of how humans interact. We present both model-based and model-free methods, and summarize the predictive performance of each method under different conditions. 
\end{abstract} 

\section{Introduction} 
The benefit of using high-fidelity simulations of a system of interest is that it produces results that closely match observations of the real-world system. 
Ideally, algorithmic design optimization would be performed using such accurate models. However, the cost of high-fidelity simulation (computational or financial) often prohibits running more than a few simulations during the optimization process. The high-fidelity models built from such a small number of high-fidelity data points tend not to generalize well. 
One approach to overcoming this limitation is to use multi-fidelity optimization, where both low-fidelity and high-fidelity data are used together during the optimization process \citep{Robinson2006}. 

When the system of interest involves humans, traditionally human-in-the-loop (HITL) experimentation has been used to build models for design optimization. A common approach in HITL experimentation is to make the simulation environment match the true environment as closely as possible \citep{Aponso2009}. A major disadvantage of this approach is that designing a good high-fidelity HITL simulation requires substantial human and technological resources. Experimentation requires highly trained participants who need to physically visit the test location. As a consequence, collecting large amounts of high-fidelity data under different conditions is often infeasible. 

In many cases, one could cheaply implement a lower fidelity simulation environment using test subjects who are not as extensively trained, allowing for the collection of large amounts of less accurate data. Indeed, this is the idea behind computational testbeds like Mechanical Turk \citep{Kittur2008, Paolacci2010}. The goal of this paper is to extend multi-fidelity concepts to HITL experimentation, allowing for the combination of plentiful low-fidelity data with more sparse high-fidelity data. This approach results in an inexpensive model of interacting humans that generalizes well to real-world scenarios.

In this paper, we begin by outlining the self-separation scenario used as a testbed to study multi-fidelity models of interacting humans. In \cref{sec:multifid} we introduce the multi-fidelity methods used in this paper. In \cref{sec:modelfree}, we present model-free approaches to multi-fidelity prediction, and in \cref{sec:modelbased} we present model-based methods. Results, summarized in \cref{sec:results}, show the benefit of incorporating low-fidelity data to make high-fidelity predictions and that model-based methods out-perform model-free methods. \Cref{sec:discussion} concludes with a summary of our findings and future directions for investigation.

\section{Modeling Interacting Humans}
\label{sec:interactinghumans}
\begin{figure}
\centering
\subfloat[\label{fig:geom}Example geometry with model variables depicted.]{
\resizebox{\columnwidth}{!} {
\begin{tikzpicture}[]
\path[use as bounding box] (-3,-4.5) grid (3.5,4.5);
\def\initrad{3.5}
\def\sinitrad{3}
\def\srad{2}
\coordinate (own) at (-90:\initrad);
\coordinate (intruder) at (100:\initrad);
\coordinate (own s init) at (-90:\sinitrad);
\coordinate (intruder s init) at (100:\sinitrad);
\coordinate (own s) at (-90:\srad);
\coordinate (intruder s) at (100:\srad);
\tikzstyle{state}=[circle,inner sep=2pt]
\tikzstyle{own state}=[state,draw=blue, fill=blue!10]
\tikzstyle{intruder state}=[state,draw=red, fill=red!10]
\draw [fill,inner color=red,draw=white] (own s) ellipse (.75 and 0.5);
\node [intruder state,xshift=-3mm,yshift=-2mm] at (own s) {};
\node [left,yshift=-5mm,xshift=-3mm] at (own s) {$o^2$};
\draw [fill,inner color=blue,draw=white] (intruder s) ellipse (.75 and 0.5);
\node [own state,xshift=-3mm,yshift=2mm] at (intruder s) {};
\node [left,yshift=3mm,xshift=-3mm] at (intruder s) {$o^1$};
\draw [thick,blue,dashed] (own s) -- (0,0);
\draw [thick,red,dashed] (intruder s) -- (0,0);
\node [aircraft top,minimum width=1cm,fill,rotate=90,xshift=-4mm] at (own s){};
\node [aircraft top,minimum width=1cm,fill,rotate=-80,xshift=-4.5mm] at (intruder s) {};
\node [draw=blue,fill=blue!10,state] at (own s) {};
\node [right] at (own s) {$s^1$};
\node [draw=red,fill=red!10,state] at (intruder s) {};
\node [right] at (intruder s) {$s^2$};
\node [own state] (s1final) at (-.5,-.5) {};
\draw [blue,->,out=100,in=-60] (own s) to (s1final) node [above, black,yshift=1mm] {$s^1_\text{f}$};
\node [intruder state] (s2final) at (.75,.75) {};
\draw [red,->,out=-80,in=170] (intruder s) to (s2final) node [right, black,yshift=2mm]  {$s^2_\text{f}$};
\node [xshift=-2mm] at ($(own s)!0.5!(s1final)$) {$a^1$};
\node [xshift=1mm] at ($(intruder s)!0.5!(s2final)$) {$a^2$};
\end{tikzpicture}
}
}\\
\subfloat[\label{fig:dag}Model structure. Shaded circles are observable variables, white circles are 
unobservable variables, squares are decision nodes, and arrows represent 
conditional relationships between variables.]{
\resizebox{\columnwidth}{!} {
\begin{tikzpicture}[>=stealth',node distance=0.5cm,column sep=3.25cm,row sep=3.25cm]
\tikzstyle{node}=[draw,circle,minimum width=1.2cm]
\tikzstyle{hidden node}=[node]
\tikzstyle{observed node}=[node,fill=gray!50]
\tikzstyle{decision node}=[node,rectangle,minimum width=1cm,minimum height=1cm]
\tikzstyle{plate}=[draw=gray,very thick,inner sep=5pt]
\matrix[ampersand replacement=\&] {
\node [node] (s1) {$s^1$}; \& \node [node] (s2) {$s^2$};\\
\node [decision node] (a1) {$a^1$}; \& \node [decision node] (a2) {$a^2$};\\[-1cm] 
\node [node] (s3) {$s^1_\text{f}$}; \& \node [node] (s4)  {$s^2_\text{f}$};\\
};
\node [observed node,above right=of a1] (oow1) {$o_\text{ow}^1$};
\node [observed node,above right=of a2] (oow2) {$o_\text{ow}^2$};
\node [observed node,right=of oow1] (oin1) {$o_\text{in}^1$};
\node [observed node,right=of oow2] (oin2) {$o_\text{in}^2$};
\draw [->]
(s1)
                edge (oow2)
                edge (a1)
                edge [in=110,out=250] (s3)
               edge (oin2)
(s2)
                edge (oow1)
                edge (a2)
                edge [in=110,out=250] (s4)
                edge (oin1)
(a1)
                edge (s3)
(a2)
                edge (s4)
(oow1)
                edge (a1)
(oow2)
                edge (a2)
(oin1)
               edge (a1)
(oin2)
                edge (a2)
;
\end{tikzpicture}
}
}
\caption{\label{fig:game}Visual self-separation game.}
\end{figure}
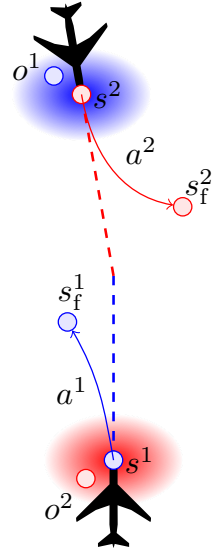
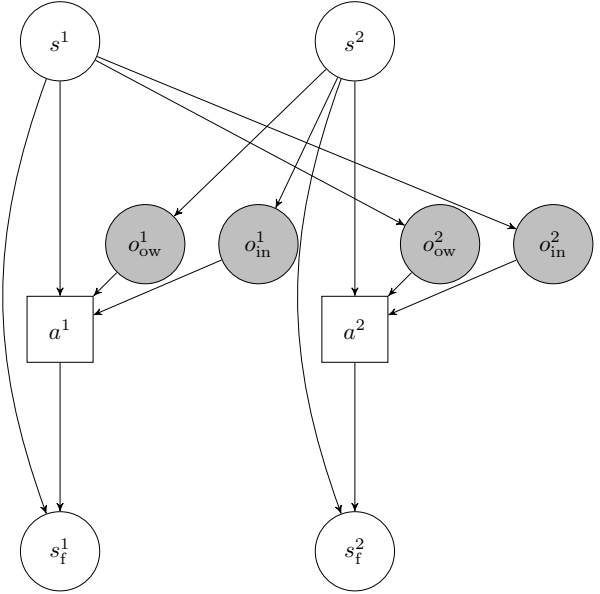

In order to ground our discussion of multi-fidelity methods in a concrete scenario, we will begin by presenting our testbed for studying the behavior of interacting humans. We model the interaction of two pilots whose aircraft are on a near-collision course. Both pilots want to avoid collision (\cref{fig:geom}), but they also want to maintain their heading. For simplicity, we model the pilots as making a single decision in the encounter. Their decisions are based on their beliefs about the other aircraft's position and velocity, their degree of preference for maintaining course, and their prediction of the other pilot's decision.

We use a combination of Bayesian networks and game-theoretic concepts \citep{Lee2010} to model this scenario. The structure of the model is shown in \cref{fig:dag}. The state of aircraft $i$ is a four-dimensional vector representing the horizontal position and velocity information. The initial distribution over states is described in \cref{sec:multifid}. The action $a^i$ taken by the pilot of aircraft $i$ represents the angular change in heading, away from their current heading. Each pilot observes the state of the other aircraft based on their view out the window $o_\text{ow}^i$ and their instrumentation $o_\text{in}^i$. Each pilot has  a weight $w^i$ in their utility function reflecting their relative desire to avoid collision versus  maintaining heading. We consider the situation where each player knows their own type (i.e., the value of the weight term in their utility function) in addition to the type of the other player \citep{Myerson1997}.

As described in this section, the pilots chose their actions based on their observations and utility function. After both pilots select an action, those actions are executed for \SI{5}{\second}, bringing the aircraft to their final states $s^i_\text{f}$.  Without loss of generality we describe the scenario from the perspective of one of the pilots, Player 1, with the other aircraft corresponding to Player 2. The model assumptions, detailed below, are symmetric between players. It should be noted that the approaches described in this paper are not limited to cases in which there are only two humans interacting, although scenarios with more humans would increase the amount of data required to make accurate predictions. 

\subsection{Visual inference}
Player 1 infers the state of the intruder based on visual information out the window and instrumentation. Player 1 is not able to exactly infer the state of Player 2 as humans are not able to perfectly estimate position and velocity based on visual information \citep{Graf2005,Schlicht2007}, and the instruments also have noise associated with their measurements (\cref{fig:dag}).

We model the out-the-window visual observation as Gaussian with a mean of the true state of the intruder aircraft and covariance matrix given by $\text{diag}(\SI{900}{\feet}, \SI{900}{\feet}, \SI{318}{\feet\per\second},  \SI{318}{\feet\per\second})$. The instrument observation is Gaussian with a mean of the true state of the intruder with covariance given by $\text{diag}(\SI{600}{\feet}, \SI{600}{\feet}, \SI{318}{\feet\per\second},  \SI{318}{\feet\per\second})$. The out-the-window velocity uncertainties are derived from the visual psychophysics literature  \citep{Weis2002,Graf2005}.

\subsection{Decision-making}
Theoretical models of multi-agent interaction offer a framework for formalizing decision-making of interacting humans. Past research has focused on game-theoretic approaches \citep{Myerson1997} and their graphical representations \citep{Koller2003,Lee2010} where agents make a single decision given a particular depth of reasoning between opponents. Such approaches have been successfully used for predicting human decision-making in competitive tasks \citep{Camerer2003,Wright2010,Wright2012}.

In cases when it is important to capture how people reason across time, sequential models of human interaction can be used \citep{Littman1994}. Such models can reflect either cooperative settings where agents are attempting to maximize mutual utility \citep{Amato2009} or competitive contexts where agents are attempting to maximize their own reward  \citep{Doshi2005,Doshi2006}. Graphical forms of sequential decision processes have also been developed \citep{Zeng2010}. 

Since humans often reason over a relatively short time horizon \citep{Sellitto2010}, especially in sudden, stressful scenarios, we model our interacting pilots scenario as a one-shot game.  Like \citet{Lee2010}, we use level-$k$ relaxed strategies as a model for human decision-making. A level-0 strategy chooses actions uniformly at random. A level-$k$ strategy assumes that the other players adopt a level-$(k-1)$ strategy \citep{Costa2001}. In our work, we assume $k=1$ because humans tend to use shallow depth-of-reasoning \citep{Wright2010}, and increasing $k>1$ had little effect on the predicted joint-decisions for our scenario. 

The focus on bounded rationality stems from observing the limitations of human decision making. Humans are unable to evaluate the probability of all outcomes with sufficient precision and often make decisions based on adequacy rather than by finding the true optimum \citep{Simon1956,Simon1982,Caplin2010}. Because decision-makers lack the ability and resources to arrive at the optimal solution, they instead apply their reasoning only after having greatly simplified the choices available. This type of sampling-based, bounded-rational view of perceptual processing has been used in computational models of human visual tracking performance \citep{Vul2010} and has been shown to predict human decision-making \citep{Vul2009}. 

\begin{figure*}
\centering
\begin{tikzpicture}[font=\scriptsize]
\begin{groupplot}[group style={group size=3 by 3,xticklabels at=edge bottom,yticklabels at=edge left,group name=plots,horizontal sep=1ex,vertical sep=1ex},axis on top,height=4cm,width=4cm,xmin=-80,xmax=80,ymin=-80,ymax=80]
\nextgroupplot[ylabel={Player 2 Decision}]
\addplot graphics [xmin=-80,xmax=80,ymin=-80,ymax=80] {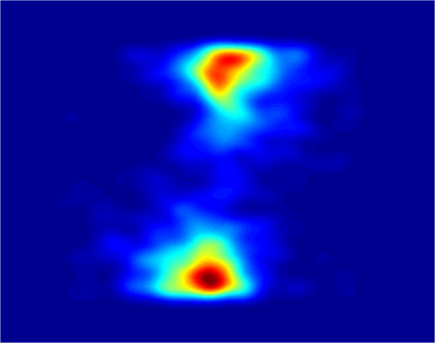};
\nextgroupplot[]
\addplot graphics [xmin=-80,xmax=80,ymin=-80,ymax=80] {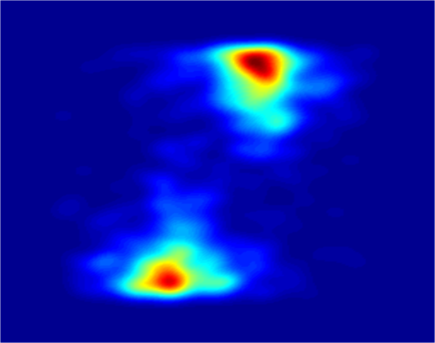};
\nextgroupplot[]
\addplot graphics [xmin=-80,xmax=80,ymin=-80,ymax=80] {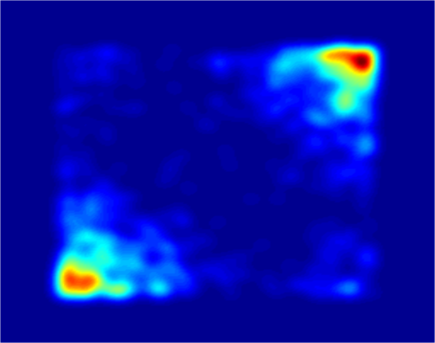};
\nextgroupplot[ylabel={Player 2 Decision}]
\addplot graphics [xmin=-80,xmax=80,ymin=-80,ymax=80] {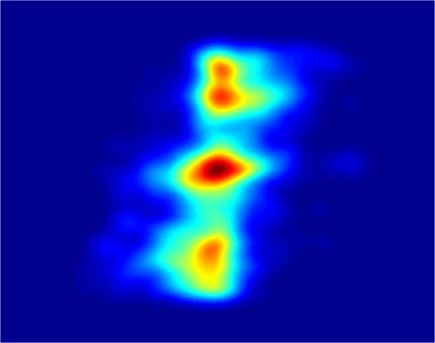};
\nextgroupplot[]
\addplot graphics [xmin=-80,xmax=80,ymin=-80,ymax=80] {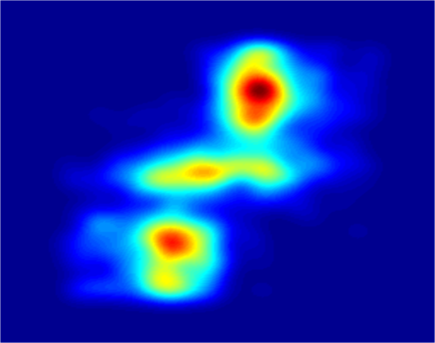};
\nextgroupplot[]
\addplot graphics [xmin=-80,xmax=80,ymin=-80,ymax=80] {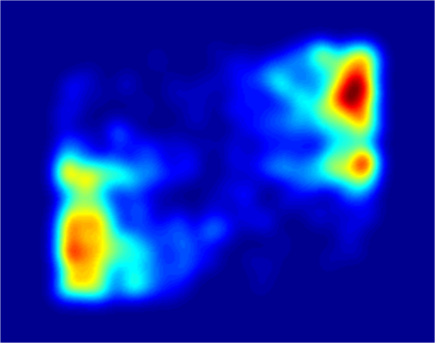};
\nextgroupplot[ylabel={Player 2 Decision},xlabel={Player 1 Decision}]
\addplot graphics [xmin=-80,xmax=80,ymin=-80,ymax=80] {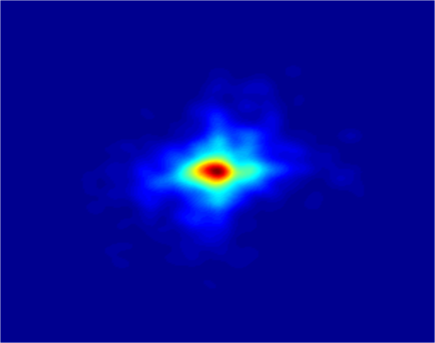};
\nextgroupplot[xlabel={Player 1 Decision}]
\addplot graphics [xmin=-80,xmax=80,ymin=-80,ymax=80] {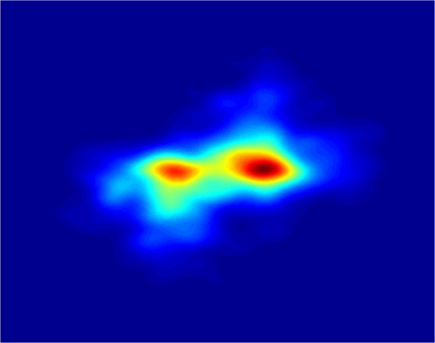};
\nextgroupplot[xlabel={Player 1 Decision}]
\addplot graphics [xmin=-80,xmax=80,ymin=-80,ymax=80] {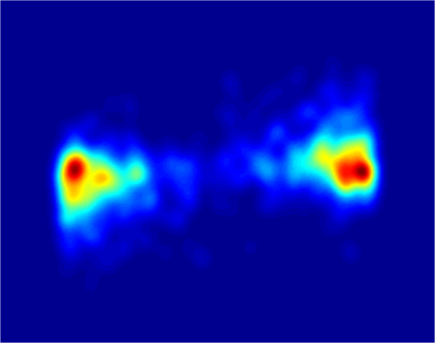};
\end{groupplot}
\node at (plots c1r1.west) [xshift=-1.5cm,anchor=east] { $0.98$};
\node at (plots c1r2.west) [xshift=-1.5cm,anchor=east] { $0.89$};
\node at (plots c1r3.west) [xshift=-1.5cm,anchor=east] { $0.80$};
\node [font=\footnotesize] at (plots c1r2.west) [xshift=-2.75 cm,rotate=90] { Player 2 Utility Weight};

\node at (plots c1r3.south) [yshift=-1cm,anchor=north] { $0.80$};
\node at (plots c2r3.south) [yshift=-1cm,anchor=north] { $0.89$};
\node at (plots c3r3.south) [yshift=-1cm,anchor=north] { $0.98$};
\node [font=\footnotesize] at (plots c2r3.south) [yshift=-1.75cm] { Player 1 Utility Weight};

\end{tikzpicture}
\caption{Influence of utility weights on joint-decision densities. Red regions depict areas with high probability joint-decisions (i.e., heading change, in degrees), while blue regions represent low probability joint-decisions.}
\label{fig:utilfig}
\end{figure*}
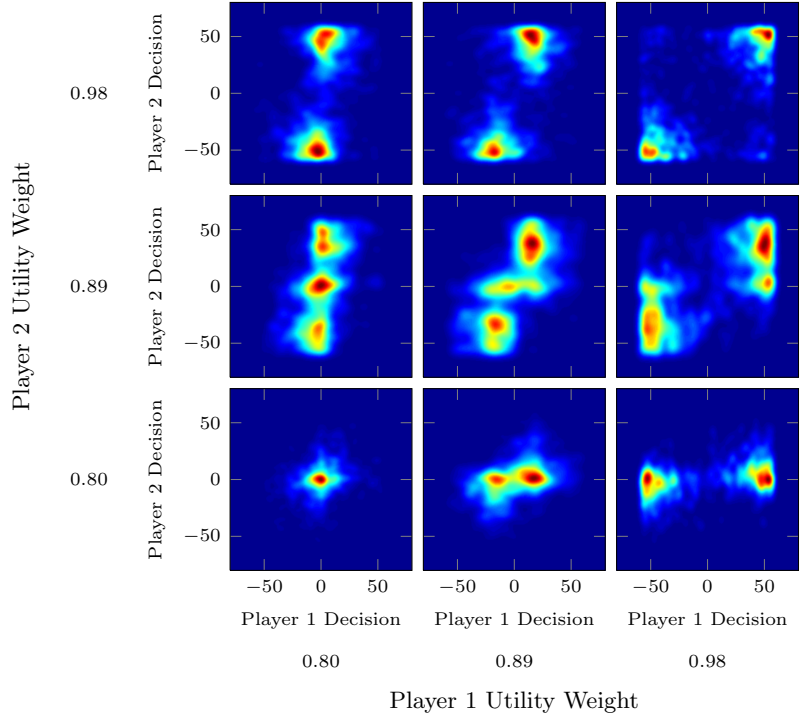

To align our model with a bounded-rational view of human performance, we assume that Player 1 makes decisions based on $m'$ sampled intruder locations and $m$ candidate actions sampled from a uniform distribution over $\pm 1 $ radian. In other words,
\begin{align}
o^{(1)}, \ldots, o^{(m')} &\sim p(o_\text{ow}^1 \mid s^2)p(o_\text{in}^1 \mid s^2)\\
a^{(1)}, \ldots, a^{(m)} &\sim \mathcal{U}(-1, 1)
\end{align}
Player 1 selects the sampled action that maximizes the expected utility over the $m'$ sampled states of the other aircraft:
\begin{eqnarray}
a = \argmax_{i} \sum_{j} w^1 d(s^1_{\text{f}\mid a^{(i)}}, s^2_{\text{f}\mid o^{(j)}})  - (1-w^1) | a^{(i)} |
\end{eqnarray} 
In the equation above, 
\begin{itemize}
\item $s^1_{\text{f}\mid a^{(i)}}$ is the final state of Player 1 after performing action $a^{(i)}$,
\item $s^2_{\text{f}\mid o^{(j)}}$ is the expected final state of Player 2 given observation $o^{(j)}$ and assuming a random heading change,
\item $d(\cdot, \cdot)$ represents Euclidean distance, and
\item $w^1$ is the relative desire of Player 1 to avoid the intruder versus maintaining their current heading.
\end{itemize}

The first term in the utility function rewards distance between aircraft, and the second term penalizes the magnitude of the heading change. \cref{fig:utilfig} shows how joint-actions change across different $w^1$ and $w^2$ combinations. When weights are relatively low (e.g., $w^1 = 0.80$, $w^2 = 0.80$), joint-actions are centered around $(0,0)$, indicating pilots tend to make very small heading change maneuvers. Conversely, when weights are relatively high (e.g., $w^1 = 0.98$, $w^2 = 0.98$), pilots tend to make large heading change maneuvers.

\subsection{Executing Actions}
\label{sec:executingactions}
After choosing an appropriate action, the pilots then execute the maneuver. Although it is known that humans have uncertainty associated with their movements \citep{Harris1998}, we simulate the case in which pilots perfectly execute the action selected by level-$k$ relaxed strategies, since uncertainty in action execution is not a focus of this study.  

Recall that the action of the player is the change in heading angle of the aircraft, away from the current heading.  From the initial state and the commanded heading change of the aircraft, the final state is obtained by simulating the aircraft for 5 seconds assuming point-mass kinematics, no acceleration, and instantaneous heading change.  

\section{Multi-fidelity Prediction}
\label{sec:multifid}
There are two ways one could distinguish a low-fidelity and high-fidelity model of this encounter. The first is using ``low-fidelity humans'' versus ``high-fidelity humans,'' where low-fidelity humans would represent people who have been trained in flight simulators but who are not commercial pilots. The second way is to have both a low-fidelity and high-fidelity simulation environment, say having low-fidelity environment being a flight simulator that does not perfectly match the conditions of an actual commercial jet. For this paper, our low-fidelity simulators are the same as the high-fidelity simulation except there is no instrument panel. The low-fidelity simulation has the $o_{\text{in}}$ nodes removed in \cref{fig:dag}.

We used separate training and testing encounters, all sampled with Player 2 being in Player 1's field of view, and a heading such that a collision occurs if an avoidance maneuver is not performed (dashed lines in \cref{fig:geom}). Training encounters were initialized with Player 2 randomly approaching at $-45^{\circ}$, $0^{\circ}$, and $+45$. Test encounters were initialized with Player 2 randomly approaching at $-22.5^{\circ}$ and $+22.5^{\circ}$. Initial headings for both train and test encounters were sampled from a Gaussian distribution with a mean around a heading direction that would result in a collision with a standard deviation of $5^{\circ}$.   

The goal of multi-fidelity prediction is  to combine  the data in the high and low-fidelity training encounters to predict the joint-decisions of pilots in the high-fidelity game. The next two sections describe different approaches to achieving this type of prediction, and the notation is summarized in \cref{tab:notation}. 

\begin{table}[h!]
\caption{\label{tab:notation}Notation used for multi-fidelity prediction}
\begin{tabular}{ll}
\toprule
& Description \\
\midrule
$A$	 		& Observed joint-actions ($a^1,a^2$) \\
$\hat{A}$  		& Predicted joint-actions  ($\hat{a}^1,\hat{a}^2$) \\
$S$		 	& Encounter geometry  ($s^1,s^2$) \\
$N$		 	& Number of encounters \\
$w$		 	& Joint utility-weights ($w^1,w^2$) \\
$z$		& Regression weight \\
$(.)^{tr}$	 	& Superscript indicating training encounters \\
$(.)^{te}$	 	& Superscript indicating test encounters \\
$(.)_{l}$	 	& Subscript indicating low-fidelity game \\
$(.)_{h}$	 	& Subscript indicating high-fidelity game \\
\bottomrule
\end{tabular}
\end{table}

\section{Model-Free Prediction Approaches}
\label{sec:modelfree}
A model-free approach is one where we do not use any knowledge of the underlying game or the decision-making process. This section discusses a traditional model-free approach to predicting joint-actions as well as a model-free multi-fidelity method.

\subsection{Locally Weighted, High-fidelity}
\label{sec:lwhifi}
The simplest approach to model-free prediction is to use the state and joint-action information from the high-fidelity training data to predict the joint-actions given the states in the testing data. Using only the high-fidelity training data, we trained a high-fidelity predictor $R_h$ that predicts joint-actions $A_h^{tr}$ given the training encounters $S_h^{tr}$.  Then, the test encounters were used as inputs to the predictor, to predict the joint-actions at the new states $\hat{A}^{te} = R_h(S^{te})$. The regression model we considered was locally weighted (LW) regression, where the predicted high-fidelity joint-decisions for a particular test encounter is the weighted combination of high-fidelity joint-decisions from the training encounters. The weights are determined by the distance between the training encounters and the test encounter:
\begin{align}
\label{eq:lwregress}
\hat{A}^{te}_i &= \sum_j^{N^{tr}_h} z_{i,j} A^{tr}_{h,j} \\
\ z_{i,j} &= \frac{ e^{-d_{i,j}}}{\sum_j  e^{-d_{i,j}}  }  
\end{align}
where $d_{i,j}$ is the standardized Euclidean distance between the state in the $i$th testing encounter and the $j$th training encounter. This approach does not use any of the low-fidelity data to make predictions, so the quality of the predictions is strictly a function of the amount of high-fidelity training data. This method represents how well one might do without taking advantage of multi-fidelity data, and serves as a baseline for comparison against the multi-fidelity approach.

\subsection{Locally Weighted, Multi-fidelity}
\label{sec:lwmulti}
Our model-free multi-fidelity method uses an approach similar to the high-fidelity method, but it also takes advantage of the low-fidelity data. We use LW regression on the low-fidelity training data to obtain a low-fidelity predictor $R_l$ that predicts joint-actions $\hat{A}_l^{tr}$ for the training encounters $S_l^{tr}$. 

We use the the low-fidelity predictor to augment the high-fidelity training data. The augmented input $(S_h^{tr}, R_l(S_h^{tr}))$ is used to train our augmented high-fidelity predictor $R_{h'}$ that predicts joint-actions based on the high-fidelity data. The test encounters $S^{te}$ were used as inputs to the augmented high-fidelity predictor to predict the joint-action for the test encounters as $\hat{A}^{te} = R_{h'}(S^{te}, R_l(S^{te}))$.  In essence, we are using the predictions from the low-fidelity predictor as features for training the high-fidelity predictor. 

\section{Model-Based Prediction Approaches}
\label{sec:modelbased}
The model-free methods described above ignore what is known about how the pilots make their  decisions. In many cases, we have a model of the process with some number of free parameters. In our case, we model the players interacting in a level-$k$ environment, and treat the utility weights $w_h$ as unknown parameters that can be learned from data, allowing the prediction of behavior in new situations. Such an approach is used in the inverse reinforcement learning literature \citep{Baker2009,Ng2000}, where the goal is to learn parameters of the utility function of a single agent from experimental data. This section outlines three model-based approaches where the data comes from experiments of varying fidelities. The first two are based on \emph{maximum a posteriori} (MAP) parameter estimates, and the third is a Bayesian approach.

\subsection{MAP High-fidelity}
\label{sec:maphifi}
To demonstrate the benefit of multi-fidelity model-based approaches, we use a baseline method that exclusively relies on high-fidelity data to make  predictions.  Since we have a model of interacting humans, we can simply use the test encounters $S^{te}$ as inputs into the model and use the resulting joint-actions as predictions. However, to make accurate predictions, we must estimate the utility weights used by the pilots. To find the MAP utility weights $w_h^{*}$, we first need to estimate $ p(A \mid  S, w)$, which is the probability of the joint-actions $A$ given encounters $S$ and weight $w$.  

As an approximation, we estimate $ p(A \mid  S, w)$ by simulating the high-fidelity game using a set of 1000 novel encounters ($S^{n}$), under different utility weight combinations. We stored the resulting joint-actions for the $j$th utility weight combination, and then estimated $p(A_h^{n} \mid  w_h^j) = \sum_S p(A_h^{n} \mid S^{n}, w_h^j)$ using kernel density estimation via diffusion \citep{Botev2010}. \cref{fig:utilfig} shows how $p(A^{n}_h \mid w_{h})$ changes across a subset of $w_h$ settings. 

We can find the MAP utility weight combination by using the joint-actions $A_h^{tr}$ we observe for our training encounters $S_h^{tr}$. For each joint-action from the training data, we find the nearest neighbor joint-action in $A_h^{n}$. We sum the log-likelihood of the nearest-neighbor joint-actions for the training encounters, under each utility weight combination. The MAP utility weight combination is defined by
\begin{equation}
\label{eq:maphifi2}
w^{*}_h = \argmax_{w_h} \;  [ \ln p(w_{h}) +  \sum_n    \ln  p(A^{n}_h \mid  w_{h}) ]
\end{equation}
where $p(w_h)$ is the prior of weight vector $w_h$.
In our experiments, we assume $p(w_{h})$ is uniform over utility weight combinations. Once we have estimated the MAP utility weights, we can simulate our high-fidelity game using the test encounters and the MAP utility weights to obtain the predicted joint-action. The predicted joint-action is obtained by generating $N_s = 10$ samples from $p(A^{te}_h \mid S^{te}_i, w^{*}_h)$ and averaging them together:
\begin{align}
A^{(1)}, \ldots, A^{(N_s)} &\sim p(A^{te}_h \mid S^{te}_i, w^{*}_h)\\
 \label{eq:maphifi1} \hat{A}^{te}_{h,i} &=  \frac{1}{N_s}  \sum_{\ell=1}^{N_s}A^{(\ell)}
\end{align} 

The amount of high-fidelity data impacts the quality of the utility-weight estimate ($w_h^*$), thereby limiting its effectiveness in situations when there is little data. The method described next overcomes this limitation by fusing both high- and low-fidelity data to find the utility weights to use for prediction. 

\subsection{MAP Multi-fidelity}
In situations when there is little high-fidelity data to estimate the utility weights, it is beneficial to fuse both low- and high-fidelity data to increase the reliability of the estimate. In this approach, the predicted high-fidelity joint-actions for test encounters ($\hat{A}^{te}$) are being computed according to \cref{eq:maphifi1}, and our MAP utility weight is still found by maximizing \cref{eq:maphifi2}. However, instead of only using the high-fidelity data to estimate the utility weights, we use both the low-fidelity and the high-fidelity data to estimate the weights. Specifically, we find the log-likelihood of the nearest-neighbor joint-action $A^{n}$ for both the low- and high-fidelity joint-actions $A^{tr}_{h}$ and $A^{tr}_{l}$. Once we have estimates for the utility weights in the decision model, we can use the test encounters $S^{te}$ as inputs to the high-fidelity model to obtain predictions $\hat{A}^{te}$, similar to what was done in \cref{eq:maphifi1}. 

Although this is a straightforward way to use both low- and high-fidelity training data,  it assumes that the decision makers in the low-fidelity and high-fidelity encounters are identical. While it is plausible that the decision makers have similar utility functions to make this a valid assumption, it could also be the case that highly trained humans have very different utility functions from lesser-trained ones, or that changes in the simulation environment also cause changes in the utility function (for example, due to the decreased sense of immersion). Therefore, we would like a method that relaxes the equal utility assumption, allowing us to make predictions for games that are different for low- and high-fidelity. 

\subsection{Bayesian Multi-fidelity}
The approaches described above use the test encounters as inputs into the high-fidelity model and sample joint-actions from $p(A^{te}_h \mid S^{te}_h, w^{*}_h)$ using the MAP utility weight estimate $w^*_h$  to compute the predicted joint-actions $\hat{A}^{te}_h$. In contrast, a Bayesian approach to multi-fidelity makes its predictions by sampling joint-actions from $p(A^{te}_{h} \mid S^{te}, w^{j}_h)$ under \emph{each} of the $j$ high-fidelity utility weight combinations, and then weighting the predicted joint-actions by the probability of the utility weights, given the low- and high-fidelity joint-actions observed from the training encounters:
\begin{equation}
\hat{A}^{te}_{h,i} = \sum_{j} p(A^{te}_{h,k} \mid S^{te}_i, w^{j}_h)  p(w^{j}_h \mid A^{tr}_l, A^{tr}_h ) 
\end{equation}
where
\begin{equation}
p(w^{j}_h \mid A^{tr}_l, A^{tr}_h )  =  p(w^{j}_h \mid A^{tr}_h) \sum_{k}{p(w^{k}_l  \mid   A^{tr}_l)p(w^{k}_l ,w^{j}_h)} \notag
\end{equation}
The probability distribution $p(w^{j}_h \mid A^{tr}_h)$ was estimated using the method outlined in Section~\ref{sec:maphifi}. The low-fidelity version of this distribution $p(w^{k}_l  \mid   A^{tr}_l)$ was estimated in a similar manner for each of the $k$ low-fidelity utility weight combinations by simulating the low-fidelity game using the same novel encounters $S^n$ and kernel density estimation via diffusion methods. 

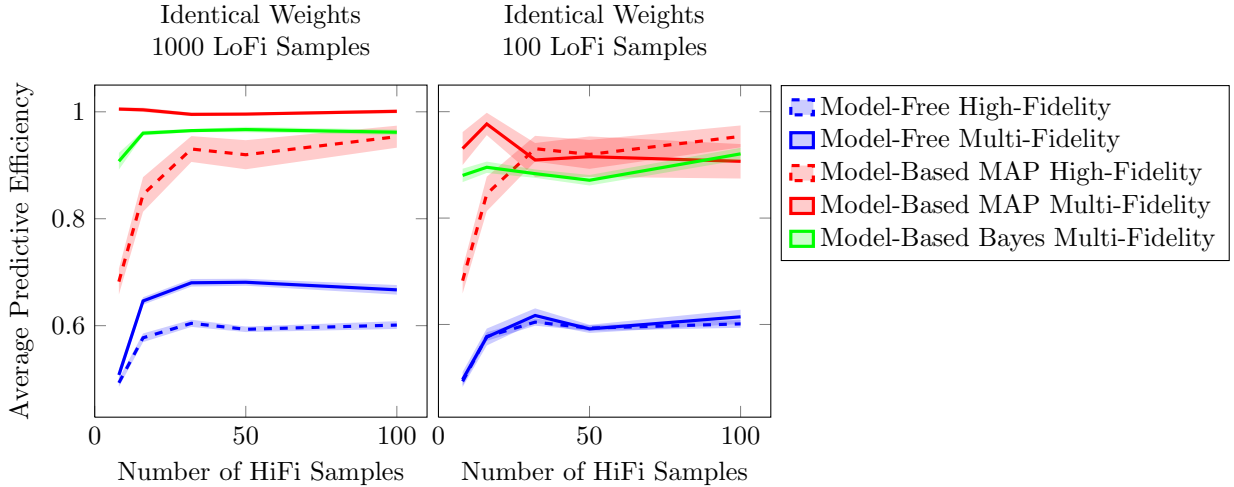
\begin{figure*}[t!]
\centering
\begin{tikzpicture}
\begin{groupplot}[group style={group size=2 by 1,xticklabels at=edge bottom,yticklabels at=edge left,group name=plots,horizontal sep=1ex},height=6cm,width=6cm,xmin=0,
        legend image code/.code={%
             \draw[very thick,#1,fill=#1,fill opacity=0.2] (0cm,-0.1cm) rectangle (0.3cm,0.1cm);
        }
]

\newcommand{\adderrorarea}[6]{
\begin{scope}[]
\pgfplotstableread{#5}\data
\pgfplotstablesave[create on use/ycoordinate/.style={create col/expr={\thisrow{#3} +\thisrow{#4}}},columns={#2,ycoordinate}
]{\data}{upper.dat}
\pgfplotstablesort[sort cmp={float >}]{\revdata}{\data}
\pgfplotstablesave[create on use/ycoordinate/.style={create col/expr={\thisrow{#3} - \thisrow{#4}}},columns={#2,ycoordinate}]{\revdata}{lower.dat}
\pgfplotstablevertcat{\filledcurve}{upper.dat}
\pgfplotstablevertcat{\filledcurve}{lower.dat}
\addplot [fill=#1,fill opacity=0.2,draw=none,forget plot] table [x=samples,y=ycoordinate] {\filledcurve};
\addplot [very thick,#1,#6] table [x=samples,y=score] {\data};
\end{scope}
}

\nextgroupplot[
xlabel={Number of HiFi Samples},
ylabel={Average Predictive Efficiency},
ymax = 1.05,
title style = {align=center},
title={Identical Weights\\1000 LoFi Samples}
]

\adderrorarea{blue}{samples}{score}{stderr}{lw_hifi_same.dat}{dashed}
\adderrorarea{blue}{samples}{score}{stderr}{lw_fused_same.dat}{}
\adderrorarea{red}{samples}{score}{stderr}{mle_hifi_same.dat}{dashed}
\adderrorarea{red}{samples}{score}{stderr}{mle_fused_same.dat}{}
\adderrorarea{green}{samples}{score}{stderr}{bayes_same.dat}{}

\nextgroupplot[
xlabel={Number of HiFi Samples},
ymax = 1.05,
title style = {align=center},
title={Identical Weights\\100 LoFi Samples},
legend pos=outer north east
]
\adderrorarea{blue}{samples}{score}{stderr}{lw_hifi_same.dat}{dashed}
\addlegendentry{Model-Free High-Fidelity}

\adderrorarea{blue}{samples}{score}{stderr}{lw_fused_same2.dat}{}
\addlegendentry{Model-Free Multi-Fidelity}

\adderrorarea{red}{samples}{score}{stderr}{mle_hifi_same.dat}{dashed}
\addlegendentry{Model-Based MAP High-Fidelity}

\adderrorarea{red}{samples}{score}{stderr}{mle_fused_same2.dat}{}
\addlegendentry{Model-Based MAP Multi-Fidelity}

\adderrorarea{green}{samples}{score}{stderr}{bayes_same2.dat}{}
\addlegendentry{Model-Based Bayes Multi-Fidelity}

\end{groupplot}
\end{tikzpicture}
\caption{Predictive performance curves for various methods across different numbers of high-fidelity training samples. Lines depict mean predictive efficiency across the $10$ simulations per sample-size condition, and the shaded-areas represent standard error. Results are shown for conditions in which the low- and high-fidelity utility weights are identical.}
\label{fig:results1}
\end{figure*}

Finally, $p(w_l ,w_h) = p(w_l^1,w^2_l,w^1_h,w^2_h)$ is a prior probability over low- and high-fidelity utility weights, and can be selected based on the degree of similarity between the high-fidelity and low-fidelity simulations. For our case, we assume a Gaussian prior with a mean corresponding to the ground-truth utility weights (\cref{sec:results}) and covariance given by:
\[
\left[\begin{array}{cccc}
.0017  & 0 & .0013 & 0  \\
0 & .0017 & 0 & .0013   \\
.0013 & 0 & .0017 & 0   \\
0 & .0013 & 0 & .0017
\end{array}\right]
\]

This covariance assumes that there is variation among players at the same fidelity, but also that the weights of the players are similar (but not identical) across fidelities. However, it assumes that the utilities of the two players within any specific encounter are independent. 

A Bayesian approach to multi-fidelity allows for predictions to be made that account for the uncertainty in how well the utility weights account for the joint-actions observed in training. Such an approach also relaxes the assumption of identical utility weights across games through a prior distribution that encodes the coupling of these parameters.  

\section{Results}
\label{sec:results}

In order to assess the performance of both model-free and model-based approaches to multi-fidelity prediction, we evaluated each method with different amounts of high-fidelity training data. We assessed the benefit of incorporating an additional 100 or 1000 low-fidelity training examples, depending on the condition.

Since novices may employ different strategies than domain experts, we also simulated the scenario under different ground-truth utility weight relationships between the low- and high-fidelity games.  For the scenario where experts and novices employ similar behavior, the ground-truth utility weights were set to: 
\begin{align}
w_l^{gt} &= \left\{{ w^1 = 0.89, w^2 = 0.90  }\right\}\\
w_h^{gt} &= \left\{{ w^1 = 0.89, w^2 = 0.90  }\right\}
\end{align}
For the scenario where there is a small difference in how novices act, the ground-truth low-fidelity utility weights were changed to:
\begin{align}
w_l^{gt} &= \left\{{ w^1_l = 0.88, w^2_l = 0.89  }\right\}
\end{align}
For the scenario where novices perform much differently than experts, the low-fidelity ground-truth utility weights were:
\begin{align}
w_l^{gt} &= \left\{{ w^1_l = 0.80, w^2_l = 0.81  }\right\}
\end{align}

\begin{figure*}[t!]
\centering
\begin{tikzpicture}
\begin{groupplot}[group style={group size=2 by 1,xticklabels at=edge bottom,yticklabels at=edge left,group name=plots,horizontal sep=1ex},height=6cm,width=6cm,xmin=0,
        legend image code/.code={%
             \draw[very thick,#1,fill=#1,fill opacity=0.2] (0cm,-0.1cm) rectangle (0.3cm,0.1cm);
        }
]

\newcommand{\adderrorarea}[6]{
\begin{scope}[]
\pgfplotstableread{#5}\data
\pgfplotstablesave[create on use/ycoordinate/.style={create col/expr={\thisrow{#3} +\thisrow{#4}}},columns={#2,ycoordinate}
]{\data}{upper.dat}
\pgfplotstablesort[sort cmp={float >}]{\revdata}{\data}
\pgfplotstablesave[create on use/ycoordinate/.style={create col/expr={\thisrow{#3} - \thisrow{#4}}},columns={#2,ycoordinate}]{\revdata}{lower.dat}
\pgfplotstablevertcat{\filledcurve}{upper.dat}
\pgfplotstablevertcat{\filledcurve}{lower.dat}
\addplot [fill=#1,fill opacity=0.2,draw=none,forget plot] table [x=samples,y=ycoordinate] {\filledcurve};
\addplot [very thick,#1,#6] table [x=samples,y=score] {\data};
\end{scope}
}

\nextgroupplot[
xlabel={Number of HiFi Samples},
ylabel={Average Predictive Efficiency},
ymax = 1.05,
title style = {align=center},
title={Small Difference in Weights\\1000 LoFi Samples}
]

\adderrorarea{blue}{samples}{score}{stderr}{lw_hifi_diff.dat}{dashed}
\adderrorarea{blue}{samples}{score}{stderr}{lw_fused_diff.dat}{}
\adderrorarea{red}{samples}{score}{stderr}{mle_hifi_diff.dat}{dashed}
\adderrorarea{red}{samples}{score}{stderr}{mle_fused_diff.dat}{}
\adderrorarea{green}{samples}{score}{stderr}{bayes_diff.dat}{}

\nextgroupplot[
xlabel={Number of HiFi Samples},
ymax = 1.05,
title style = {align=center},
title={Large Difference in Weights\\1000 LoFi Samples},
legend pos=outer north east
]
\adderrorarea{blue}{samples}{score}{stderr}{lw_hifi_diff.dat}{dashed}
\addlegendentry{Model-Free High-Fidelity}

\adderrorarea{blue}{samples}{score}{stderr}{lw_fused_diff2.dat}{}
\addlegendentry{Model-Free Multi-Fidelity}

\adderrorarea{red}{samples}{score}{stderr}{mle_hifi_diff.dat}{dashed}
\addlegendentry{Model-Based MAP High-Fidelity}

\adderrorarea{red}{samples}{score}{stderr}{mle_fused_diff2.dat}{}
\addlegendentry{Model-Based MAP Multi-Fidelity}

\adderrorarea{green}{samples}{score}{stderr}{bayes_diff2.dat}{}
\addlegendentry{Model-Based Bayes Multi-Fidelity}

\end{groupplot}
\end{tikzpicture}
\caption{Predictive performance curves for various methods across different numbers of high-fidelity training samples. Lines depict mean predictive efficiency across the $10$ simulations per sample-size condition, and the shaded-areas represent standard error. Results are shown for conditions in which the low- and high-fidelity utility weights have small (left) and large (right) differences between them.}
\label{fig:results2}
\end{figure*}
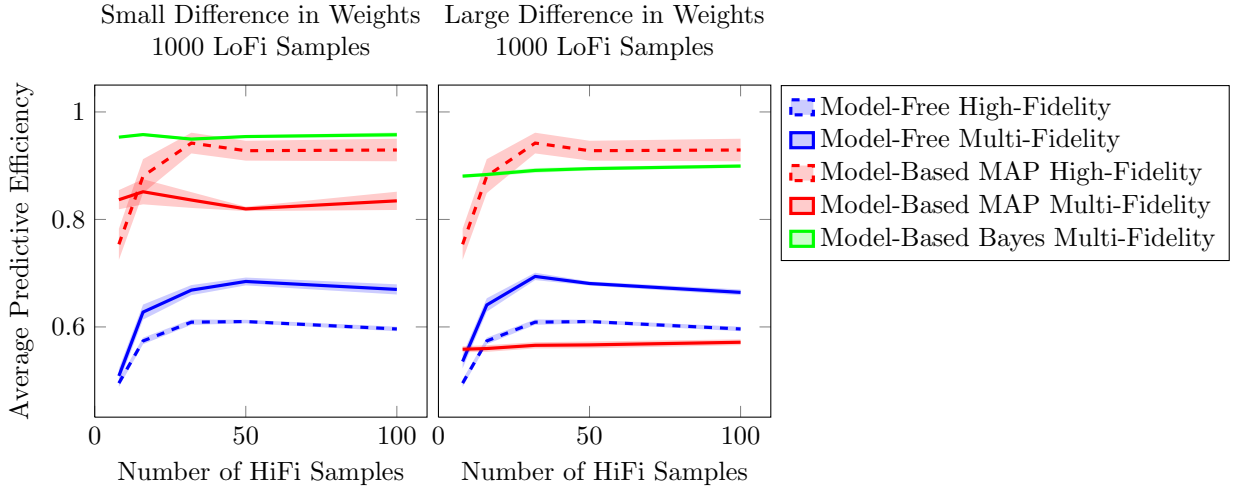

Performance is measured in terms of ``predictive efficiency,'' which is a number typically between 0 and 1. It is possible to obtain a number above 1 because the normalization factor is an estimated lower-bound of the test-set error. The test-set error is given by
\begin{equation}
D = \sum_i d(\hat A_i, A_i^{te}),
\end{equation}
where $d(\cdot, \cdot)$ is the Euclidean distance, $\hat A_i$ is the predicted action for the $i$th encounter, and $A_i^{te}$ is th-e joint-action of the $i$th test encounter. A lower-bound on the test-set error $D^{lb}$ can be approximated by estimating the error when using the ground-truth model in conjunction with \cref{eq:maphifi1}. The predictive efficiency is given by $D^{lb}/D$.

\Cref{fig:results1} shows how average predictive efficiency changes as a function of the amount of low- and high-fidelity training data for the identical utility weight condition. \Cref{fig:results2} shows how average predictive efficiency is impacted by the level of discrepancy between low- and high-fidelity utility weights.  

Multi-fidelity model-free approaches (blue solid lines) predicted better than methods that used only high-fidelity data  (blue dashed lines), except for the case in which there was little (100 samples) low-fidelity available to train the model (\Cref{fig:results1}). 

Model-based multi-fidelity approaches had better predictive performance (red and green solid lines) than high-fidelity methods (red dashed line) in cases when there was a large amount of low-fidelity data (1000 samples), and there was little or no difference in the utility weights used by the low-fidelity and high-fidelity humans in the task (\Cref{fig:results1,fig:results2}).

\section{Conclusions and Further Work} 
\label{sec:discussion}
We developed a multi-fidelity method for predicting the decisions of interacting humans.  We investigated the conditions under which these methods produce better performance than  exclusively relying on high-fidelity data. In general, our results suggest that multi-fidelity methods provide benefit if there is a sufficient amount of low-fidelity training data available and the differences between expert and novice behavior is not too large. Future effort will investigate these distinctions in greater detail.    

This approach can also be extended to many domains beyond aviation. For example, work is currently being conducted that uses a `power-grid game' to produce low-fidelity data to train algorithms that detect attacks on the system. This data could be combined with relatively sparse high-fidelity data (e.g., historical data involving known attacks) to increase the predictive performance of the model. 


\subsubsection*{Acknowledgments}
The Lincoln Laboratory portion of this work was sponsored by the Federal Aviation Administration under Air Force Contract \#FA8721-05-C-0002.  The NASA portion of this work was also sponsored by the Federal Aviation Administration. Opinions, interpretations, conclusions and recommendations are those of the author and are not necessarily endorsed by the United States Government. We would like to thank James Chryssanthacopoulos of MIT Lincoln Laboratory for his comments on early versions of this work. 
\bibliographystyle{apalike}
\bibliography{references}
\end{document}